\documentclass[3p]{elsarticle}
\usepackage[utf8]{inputenc}
\usepackage[T1]{fontenc}
\usepackage{lmodern}
\usepackage[english]{babel}
\usepackage{xcolor}
\usepackage{graphicx}
\usepackage{subcaption}
\usepackage{tikz}
\usepackage{amsmath}
\usepackage{multirow,makecell}
\usepackage{booktabs}         
\DeclareUnicodeCharacter{202F}{\,} 
\usepackage{url}

\usepackage{ulem} 

\journal{Artificial Intelligence in Medicine}

\begin{document}

\begin{frontmatter}

	\title{Deep‑Learning Atlas Registration for Melanoma Brain Metastases: Preserving Pathology While Enabling Cohort‑Level Analyses}

	\author[1]{Nanna E. Wielenberg\corref{cor1}}
	\author[1]{Ilinca Popp}
	\author[3]{Oliver Blanck}
	\author[4]{Lucas Zander}
	\author[4]{Jan C. Peeken}
	\author[4]{Stephanie E. Combs}
	\author[1]{Anca-Ligia Grosu}
	\author[2]{Dimos Baltas}
	\author[2]{Tobias Fechter}

	\affiliation[1]{organization={Department of Radiation Oncology, Medical Center – University of Freiburg, Faculty of Medicine, University of Freiburg},city={Freiburg}, country={Germany}}
	\affiliation[2]{organization={Division of Medical Physics, Department of Radiation Oncology, Medical Center – University of Freiburg, Faculty of Medicine, University of Freiburg}, city={Freiburg}, country={Germany}}
	\affiliation[3]{organization={Department of Radiation Oncology, University Medical Center Schleswig-Holstein}, city={Kiel}, country={Germany}}
	\affiliation[4]{organization={Department of Radiation Oncology, TUM School of Medicine and Health, TUM University Hospital Rechts der Isar, Technical University of Munich}, city={Munich}, country={Germany}}
	\cortext[cor1]{Corresponding author}

	\begin{abstract}

		\textbf{Purpose:}
		Melanoma brain metastases (MBM) are common, heterogeneous lesions that are difficult to compare across patients because of anatomical variability, geometric distortions, and differing MRI acquisition parameters, which hamper reliable cohort‑level analysis. To address this, we propose a deep‑learning‑based deformable registration framework that aligns the individual anatomy of brain metastases to a common anatomical atlas while preserving pathological tissue without a corresponding atlas counterpart. This enables accurate and reproducible spatial analyses of large, multi‑centre cohorts of patients with MBM.

		\textbf{Methods:}
		We present a fully differentiable, preprocessing-free atlas-registration framework. Missing correspondences caused by metastatic lesions are addressed without explicit lesion masks through a forward-model similarity evaluation on distance-transformed anatomical labels, combined with a volume-preserving loss. Registration accuracy was assessed using the Dice coefficient (DC), Hausdorff distance (HD), average symmetric surface distance (ASSD), and Jacobian-based deformation plausibility. The framework was applied to a cohort of 209 patients with melanoma brain metastases (MBM) from three centres, enabling standardized mapping of metastasis locations to anatomical, arterial, and perfusion atlases.

		\textbf{Results:}
		Models achieved excellent registration performance across all clinical datasets (DSC 0.89–0.92, HD 6.79–7.60~mm, ASSD 0.63–0.77~mm) while preserving metastatic volumes.
		Spatial analysis revealed a significant over-representation of MBM in the cerebral cortex and putamen, and an under-representation in white matter.
		Metastases were consistently located in close proximity to the gray--white matter junction, confirming a well-established spatial predilection of MBM. No arterial territory showed a significantly increased frequency of metastases after correcting for regional volume.

		\textbf{Conclusion:}
		The proposed framework enables accurate and robust atlas registration of pathological brain images without the need for preprocessing or lesion masks and supports reproducible cohort-level analyses.
		Applied to multi-centre data on melanoma brain metastases,
		it confirms and refines known spatial patterns, demonstrating preferential seeding near the gray-white matter junction
		and over-representation in cortical regions.
		The publicly available implementation of the framework provides a valuable tool for reproducible studies and facilitates future extensions to other tumour types and pathological brain conditions.
	\end{abstract}




\end{frontmatter}

\section{Introduction}

Brain metastases are a common complication of systemic malignancies and represent a major cause of morbidity and mortality in oncological patients, most frequently originating from lung cancer, breast cancer, and malignant melanoma. While lung and breast cancers account for the majority of cases due to their high overall incidence, malignant melanoma exhibits a particularly high propensity for cerebral metastasis on a per-patient basis \cite{Davies2010,Bertolini2015,Cardinal2021}.

Advances in imaging, local treatments, and systemic therapies have substantially improved the historically poor prognosis of melanoma brain metastases (MBM) and enhanced long-term disease control \cite{Fife2004,Schadendorf2018}. As patient survival has increased and neuroimaging has become more sensitive, brain metastases are diagnosed more frequently in the disease course, leading to a rising observed incidence of MBM and rendering it an increasingly relevant and clinically urgent field of research.

MBM can lead to substantial neurologic morbidity, and more than half of affected patients die as a result of central nervous system progression \cite{Eroglu2022}. Despite substantial therapeutic progress and meaningful survival improvements, the clinical management of MBM remains challenging, and outcomes for many patients remain poor. Major uncertainties persist regarding the biological behavior, treatment response, and mechanisms of recurrence of MBM, largely due to a persistent lack of high-quality data, which prevents individualized treatment decisions in the clinical routine \cite{sperduto2023}.

From an imaging perspective, the systematic analysis of brain metastases across patient cohorts is challenged by substantial anatomical variability, geometric distortions, and heterogeneous image acquisition protocols, which hinder consistent identification, comparison, and localization of anatomical structures and pathological findings. These limitations render robust inference from single-patient analyses difficult. Atlas-based image registration addresses these challenges by mapping individual scans into a common anatomical reference space, thereby reducing inter-individual variability and enabling standardized localization, automated segmentation, and reproducible cohort-level analyses, including large-scale data-driven approaches \cite{CHEN2025103385}.

The mapping that aligns each individual patient image with a common anatomical reference image (the atlas) can be either calculated by rigid, affine, deformable registration or by a combination of them. The difference lies in the degrees of freedom the transformations have and what kinds of shape change they can represent. Rigid and affine registrations can be represented by a transformation matrix, whereas deformable registration is expressed by a deformation vector field (\textit{DVF}). The deformation field specifies a three-dimensional displacement that maps, depending on the direction of the registration, every voxel in the patient image into the atlas space or every voxel in the atlas image into the patient image space. Anatomical plausible transformations should preserve spatial relationships between structures and should be invertible to transfer information in both directions.

Atlas construction approaches can be divided into backward models and forward models \cite{ding2022aladdin}. Backward models transform the images and measure the similarity between images and the atlas in the atlas space, while forward models measure the similarity in image space by transforming the atlas. Backward models are usually used when analysing properties or variations of entire populations in a common reference space, while forward models are typically applied when information from an atlas is transferred to individual cases (e.g.\ in atlas-based segmentation) \cite{BachCuadra2015}. A comprehensive overview of recent methodological advances in atlas-based image registration is provided by Chen et al. \cite{CHEN2025103385}. In the following, we will focus on related articles addressing the analysis of tumour characteristics using atlas-based methods in the brain region, especially MBM.


A key challenge when registering images of cancer patients to an atlas is that the tumour has no corresponding region in the atlas image. Without dedicated consideration of the tumour in the cost function, the tumour can get collapsed or distorted when registering the image with tumour to the atlas \cite{dong2023preserving,Brett2001}. To date, various approaches have been proposed to address this problems. One of the earliest approaches proposed masking the cost function in regions affected by focal lesions \cite{Brett2001}. Methods presented since then follow similar ideas.
Inverse consistency of bidirectional deformation fields has been analysed as a means to identify missing correspondences in deformable registration \cite{10.1007/978-3-031-33842-7_20,10.1007/978-3-031-33842-7_21}. Dong et al. trained a baseline registration model on healthy patients and apply it to cancerous patients \cite{dong2023preserving}. When the network detects disproportionate volume shrinkage, this indicates the presence of a tumour, and a soft tumour mask is generated. The soft tumour mask is then used to calculate a volume preserving loss when training a network with tumour patients.

A common limitation of these methods is that detecting missing correspondences requires thresholding, which must be tuned and can be task- or dataset-dependent. In the work by Joshi et al. the morphological change is modelled by allowing the network to change image intensities \cite{10.1007/978-3-031-45087-7_17}. However, the approach requires a mask of the pathological region to avoid that the transformation from moving to source image is modelled solely by changing voxel intensities and not deformation vectors.

Andresen et al. use a separate decoder to mask regions with missing correspondences in the registration process \cite{Andresen2022}.
To stabilize mask generation the part of the network that calculates the vector field needs to be pre-trained.
Additional approaches rely on anatomical landmarks to guide the registration process \cite{10.1007/978-3-031-33842-7_22,10.1007/978-3-031-33842-7_24,10.1007/978-3-031-44153-0_2,10.1007/978-3-031-44153-0_1}. The difficulty of aligning brain images with missing correspondencies has also been showcased in the Brain Tumour Sequence Registration Challenge \cite{Baheti2021b}, where some of the outlined techniques were assessed.

Once images are reliably aligned, it is feasible to systematically analyse tumour locations and properties at the cohort level. Previous work has demonstrated that incorporating spatial information can improve tumour grading and outcome prediction models \cite{Liu2023a}. Atlas-based analyses have further enabled unbiased and reproducible assessment of tumour location and functional region involvement, for example in lower-grade gliomas \cite{GomezVecchio2021}.
Similarly, atlas-based frameworks have been applied to investigate recurrence patterns, such as prostate bed recurrence using $^{68}$Ga-PSMA PET/CT imaging \cite{Sonni2023}.

Atlas-based registration methods provide a suitable framework for standardized, cohort-level analyses to investigate the spatial distribution of brain metastases. Specific for melanoma, a systematic review showed that metastases occur more frequently in the frontal and temporal lobes and less often in the cerebellum \cite{Cardinal2021}.
A pathology-based study analysing neuroanatomical localisation reported a predominance of frontal lobe involvement for melanoma brain metastases \cite{Bonert2023}. Furthermore, an atlas-based analysis of 13{,}067 brain metastases (409 MBM) demonstrated that metastases predominantly localize to high perfusion regions near the gray--white matter junction \cite{Barrios2025}. Specific for MBM, an increased frequency was observed in regions supplied by the anterior circulation and an under-represented presence in the cerebellum \cite{Barrios2025}. 
More recent work reported that MBM are more evenly distributed, with a modest preference for the occipital, temporal, and frontal lobes as well as regions supplied by the middle cerebral artery \cite{Lyu2025}. 

Although dedicated registration methods for atlas-based analysis and handling missing correspondences have been proposed, clinical studies have largely not adopted these approaches and instead relied on general-purpose registration tools such as ANTs \cite{Tustison2021,Lyu2025},
commercial software packages (e.g.\ MIM Software Inc., Cleveland, OH, USA) \cite{Barrios2025}, or did not employ an atlas framework \cite{Bonert2023}. Possible reasons for this limited clinical adoption include the need for extensive preprocessing, lesion masks, threshold tuning, or pre-trained auxiliary components.

\section{Contribution}
In this work, we present a deep learning-based registration framework designed for atlas registration of pathological brain images while remaining applicable beyond this specific setting. It (i) operates directly on native medical images without preprocessing, (ii) addresses missing correspondences in a simple yet effective way without lesion masks,
and (iii) enables reproducible spatial analyses. The complete framework is publicly available, and its capabilities are demonstrated through an atlas-based analysis of MBM from three independent centres. Our results confirm previously reported findings while providing additional, refined insights into the spatial distribution of MBM.


\section{Material and Methods}

Let $\mathcal{I} = \{ I_i \mid i = 1, \ldots, N \}$ be a set of $N$ three dimensional (\textit{3D}) medical image datasets and $\mathcal{C} = \{ C_i \mid i = 1, \ldots, N \}$ a set of label maps that define structures in the corresponding images. The aim of the registration framework is to find for each $I_i \in \mathcal{I}$ and $C_i \in \mathcal{C}$ a transformation $T_i \in \mathcal{T}$ that maps $I_i$ and $C_i$ to a 3D atlas $I_A$ or its label map $C_A$. Each $T_i$ is a dense vector field $u$ with 3 components (x-, y- and z-direction) which describe the trajectory for each voxel in $I_i$. The deformation vector at a given position $x$ and image $i$ is indicated as $u_i(x)$. $T_i(I_i)$ maps $I_i$ to $I_A$ by applying $T_i$ to image $I_i$ and $T_i^{-1}(I_A)$ maps $I_A$ to $I_i$.

\subsection{Datasets}
Seven different datasets were used in this work. First, a publicly available dataset from Task~3 of the 2021 Learn2Reg challenge
\cite{10.1007/978-3-030-78191-0_1,6795067}, consisting of healthy subjects, was used to develop and calibrate the registration algorithm, establish a baseline model, and make the results comparable to those of other published methods.

In addition, three atlases were used to analyze MBM characteristics within a unified reference space.

\begin{itemize}
	\item \textbf{ICBM152 Atlas} The 2020 ICBM 152 extended nonlinear symmetric atlas (\textit{ICBM152}) \cite{FONOV2011313, FONOV2009S102}, provides a sharp average T1-weighted image and extended z-direction coverage, including the lower parts of the head.
	\item \textbf{Arterial atlas} \cite{Liu2023b}: Contains 32 arterial territories segmented on ICBM152.
	\item \textbf{Perfusion atlas} \cite{Barrios2025}: Normalized perfusion map registered to ICBM152 space.
\end{itemize}

Three additional datasets (\textit{Clin1--3}) comprising patients with MBM were employed for clinical analysis. These datasets contain contrast-enhanced 3D-T1-weighted MRI scans from 209 patients with melanoma brain metastases and were collected across three independent German centres. The initial tumour contours for the \textit{Clin1--3} datasets, representing the gross tumour volume (GTV), were manually delineated by multiple radiation oncologists for radiotherapy planning. All contours were reviewed and verified by an experienced radiation oncologist prior to analysis. 

The Learn2Reg datasets comes with delineated brain structures. The contours were created with the SAMSEG tool \cite{PUONTI2016235} of the FreeSurfer framework. For consistency we delineated the healthy brain structures of the other datasets with the same tool. For each \textit{Clin1--3} and Learn2Reg case as well as the ICBM152 atlas an anatomical contour set  with 35 structures (\textit{SegAnat}) was created to steer the registration process and for detailed anatomical analysis. The arterial atlas (\textit{SegArt}) consists of 32 arterial territories, whilst the perfusion atlas is a normalized perfusion map. The image properties of the used datasets are given in Table \ref{tab:datasetDetails} and segmented regions are listed in Table \ref{tab:structureSetDetails}. 

\begin{table*}[ht!]
	\centering
	\caption{Image details of the used datasets}
	\label{tab:datasetDetails}
	\begin{tabular}{lcccl}
		\hline
		\textbf{Dataset} & \textbf{Cases} & \textbf{Voxel Size (mm)}          & \textbf{Image Matrix}        & \textbf{Notes} \\
		\hline
		Learn2Reg        & 414            & 1.00~x~1.00~x~1.00                & 256~x~256~x~256              & Healthy        \\
		Clin1            & 112            & 0.49-1.00~x~0.49-1.00~x~0.90-2.00 & 256-512~x~256-512~x~95-192   & MBM            \\
		Clin2            & 78             & 0.60-0.92~x~0.60-0.92~x~1.00      & 512~x~512~x~229-378          & MBM            \\
		Clin3            & 19             & 0.23-1.00~x~0.23-1.00~x~0.75-2.51 & 230-1024~x~240-1024~x~59-267 & MBM            \\
		ICBM152          & 1              & 1.00~x~1.00~x~1.00                & 193~x~239~x~263              & Atlas          \\
		\makecell[l]{Perfusion                                                                                                \\Atlas} & 1 & 1.00~x~1.00~x~1.00 & 193~x~239~x~263 & Atlas\\
		\hline
	\end{tabular}

\end{table*}

\begin{table*}[ht!]
	\centering
	\caption{Structure sets of the anatomical (\textit{SegAnat}) and the arterial (\textit{SegArt}) atlases used in the experiments.}
	\label{tab:structureSetDetails}
	\begin{tabular}{l p{12cm}}
		\hline
		Structure Set & Containing Structures                                                                                                                                                                                                                                                                                                                                                                                                                                                                                                                                                                                                                                                                                                                     \\
		\hline
		SegAnat       & Background, Cerebral White Matter Left/Right, Cerebral Cortex Left/Right, Lateral Ventricle Left/Right, Inferior Lateral Ventricle Left/Right, Cerebellum White Matter Left/Right, Cerebellum Cortex Left/Right, Thalamus Left/Right, Caudate Left/Right, Putamen Left/Right, Pallidum Left/Right, 3rd Ventricle, 4th Ventricle, Brain Stem, Hippocampus Left/Right, Amygdala Left/Right, Accumbens Area Left/Right, Ventral Diencephalon Left/Right, Vessel Left/Right, Choroid Plexus                                                                                                                                                                                                                                                   \\
		\hline
		SegArt        & Background, Anterior Cerebral Artery Left/Right, Medial Lenticulostriate Left/Right, Lateral Lenticulostriate Left/Right, Frontal Pars of Middle Cerebral Artery Left/Right, Parietal Pars of Middle Cerebral Artery Left/Right, Temporal Pars of Middle Cerebral Artery Left/Right, Occipital Pars of Middle Cerebral Artery Left/Right, Insular Pars of Middle Cerebral Artery Left/Right, Temporal Pars of Posterior Cerebral Artery Left/Right, Occipital Pars of Posterior Cerebral Artery Left/Right, Posterior Choroidal and Thalamoperfurators Left/Right, Anterior Choroidal and Thalamoperfurators Left/Right, Basilar Left/Right, Superior Cerebellar Left/Right, Inferior Cerebellar Left/Right, Lateral Ventricle Left/Right \\
		\hline
	\end{tabular}
\end{table*}

\subsection{Preprocessing}

The proposed atlas registration framework requires images and atlas to be roughly pre-aligned. This was done by using the affine registration methods of FLIRT v6.0 \cite{JENKINSON2001143,JENKINSON2002825} (L2R) and the general registration module \cite{JohnsonHarrisWilliams2007} of 3D Slicer (v.\ 5.6.0) \cite{FEDOROV20121323}. For each dataset, both methods were applied, and the registration matrix that yielded the higher mean Dice coefficient for all $C_i \in \mathcal{C}$ and $C_A$ was stored for subsequent processing. The affine pre-alignment is used solely for coarse initialization and does and does not involve any explicit image preprocessing or modification of pathological regions, such as skull stripping, intensity normalization, or lesion masking.

\subsection{Network Architecture}

\begin{figure*}[htb]
	\centering
	\mbox{
		\includegraphics[width=1.0\linewidth]{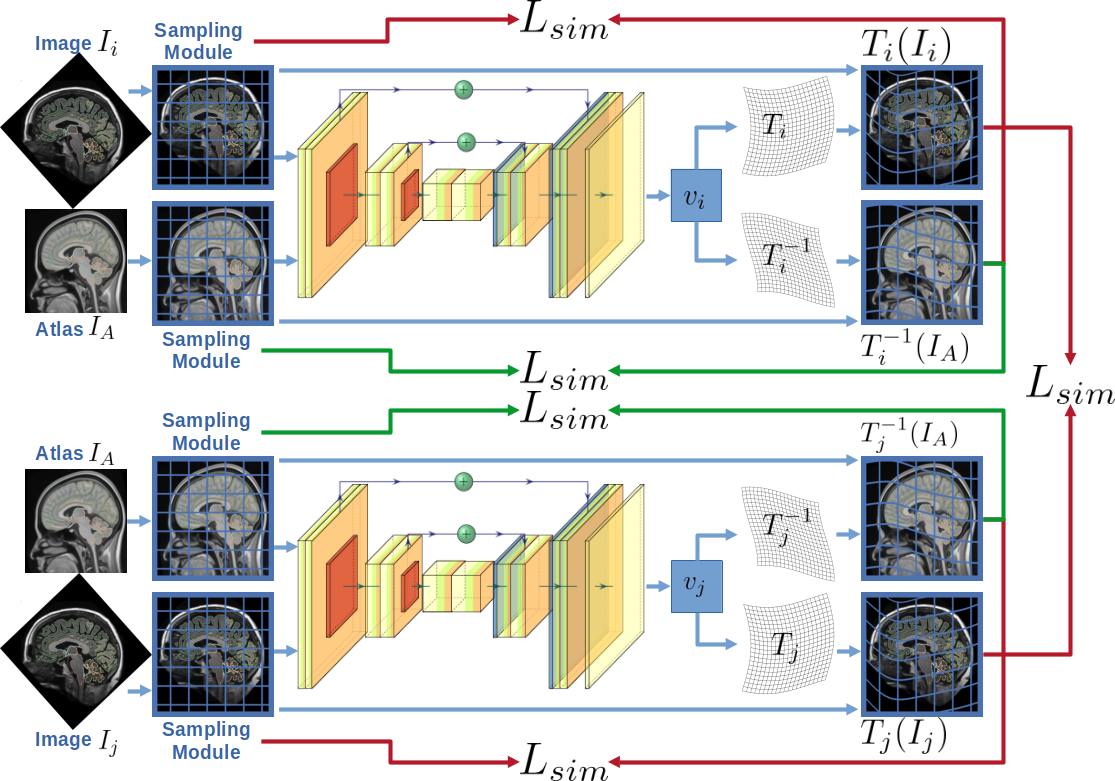}
	}
	\caption{The overall architecture used in this work. The sampling modules act like a wrapper around images ($I_i$ and $I_j$) and the atlas ($I_A$) and are responsible for preparing them as input for the neural network and for applying the calculated transformations ($T_i$, $T_j$) and their inverse transformations ($T_i^{-1}$, $T_j^{-1}$). The neural network (here a U-net) calculates a stationary velocity field $v$, which is integrated into the final transformations $T_i$ and $T_i^{-1}$ by scaling and squaring. Blue arrows indicate data flow, while red and green arrows show which images are used for the similarity loss $L_{sim}$
	of the general and the over-fitted model, respectively.
	Regularization losses are not shown in this figure.}
	\label{fig:networkArchitecture}
\end{figure*}

The proposed architecture consists of two main parts:
a sampling module that prepares and standardizes the input,
and a deep neural network that calculates the DVF. The whole architecture is depicted in Figure~\ref{fig:networkArchitecture}.

\subsubsection{Sampling Module}

Medical images are not standardized and can come with various image properties. 
However, for training a neural network, standardized input images are required. The standardization is usually done in a preprocessing step by resampling and interpolation operators, which is connected with a degradation of image quality and a loss of information \cite{MURRAY2024108422}. Applying a transformation $T_i$ to an image $I_i$ again requires interpolation and can amplify image deterioration. To mitigate this issue, we use a sampling module that renders resampling, standardization, and normalization in a preprocessing step unnecessary. The module stores a sampling grid for each image $I_i$ and can be seen as an image wrapper. The properties (e.g. size and spacing) of the grid define the output image. When a transformation is applied to an image $T_i(I_i)$, it is not the image itself that is transformed, but rather the sampling grid. The original image remains unchanged and is sampled according to the transformed grid. This approach reduces image degradation caused by interpolation.

\subsubsection{Neural Network Architecture}

For this work, we used a U-net architecture \cite{RFB15a} with 2 downsampling steps with average pooling layers, 2 upsampling steps with transposed convolution layers \cite{dumoulin2018guideconvolutionarithmeticdeep},
skip connections by summation instead of concatenation, and a scaling and squaring \cite{Arsigny2006} layer. The network takes an image $I_i$ and an atlas $A$ as input and calculates a stationary velocity field $v_i$.
The final transformation $T_i$ and $T_i^{-1}$ are obtained by integrating $v_i$ and $-v_i$ through scaling and squaring.
The U-net was chosen due to its simple design and good performance, but in the proposed implementation it can be easily replaced by more complex architectures.


\subsection{Training Procedure}

In this study, we made use of two different training steps. First, a general model was optimized on a set of training datasets. In a second step at inference, the model obtained in step one was over-fitted in a one-shot manner \cite{Fechter2020} to the image to be registered. The general model was trained in a forward approach, while over-fitting was done with a backward approach. In the forward approach, evaluation is performed in image space, which has the advantage that extra considerations for missing correspondences are not necessary (see Section~\ref{sec:missingCorrespondences}). In the over-fitting step, we used a backward approach to achieve the highest possible accuracy in the atlas space. 
The general model was trained for 350 epochs and the over-fitting model for 1500 epochs.

\subsection{Loss Function}


As mentioned before, we used two different loss functions that differ in the considered direction of transformation and in the inclusion of a volume preserving loss. For the general model, the loss was defined as:

\begin{equation}
	\label{eqn:lossGeneral}
	L_{G} = \sum_{i=1}^{N} \lambda_1 L_{sim}\big(C_i,T_i^{-1}(C_A)\big) + \lambda_2 L_{reg}\big(T_i\big) + \lambda_3 \sum_{\substack{j=1 \\ j \neq i}}^{N} L_{sim}\big(T_i(C_i), T_j(C_j)\big),
\end{equation}

whereas the over-fitting loss was:

\begin{equation}
	\label{eqn:lossOverfitting}
	L_{OF} = \sum_{i=1}^{N} \lambda_1 L_{sim}\big(T_i(C_i),C_A\big) + \lambda_2 L_{reg}\big(T_i\big) + \lambda_4 L_{vol}\big(C_A, T_i\big).
\end{equation}

$L_{sim}$ measures the dissimilarity between two volumes, $L_{reg}$ promotes smooth transformations, and $L_{vol}$ encourages the volume preservation of anatomical structures.
The first $L_{sim}$ term in Equations~\ref{eqn:lossGeneral} and~\ref{eqn:lossOverfitting} ensures that images and atlas align well.
The second $L_{sim}$ term in Equation~\ref{eqn:lossGeneral} ensures that images align well in atlas space~\cite{ding2022aladdin}. Equation~\ref{eqn:lossOverfitting} has no second $L_{sim}$ term because only one image was registered to the atlas during over-fitting. Measuring $L_{sim}$ by comparing voxel values of atlas and images can be affected by the inherent blurriness of the atlas, image artefacts, and heterogeneous image properties. Therefore, we decided to use label maps defining organ boundaries as a surrogate. However, the overlap of label maps provides for each voxel only a binary value: either a label match or a mismatch. To obtain a continuous measure with spatial depth information, we converted each label map to a distance map $D(C_i)$
and calculated the negative normalized cross correlation (\textit{NCC}) between label maps:

\begin{equation}
	\label{eqn:lossSim}
	L_{sim}(C_i, C_j) = -\operatorname{NCC} \big(D(C_i),D(C_j)\big).
\end{equation}

The value of $D(C_i)$ at position $x$ is defined as:
\begin{equation}
	\label{eqn:distanceMaps}
	D(C_i)(x) = \min_{c_j \in \mathcal{U}(C_i)} \left( \max\left(0, \operatorname{dist} \big( x, c_j \big) \right) \right) + \gamma C_i(x),
\end{equation}
where $\operatorname{dist} \big(x, c_j\big)$ denotes the signed distance from $x$ to the boundary of label $c_j$ (with positive values inside and negative values outside of the contour), the $\max\left(\cdot\right)$ operator ensures that only labels that contain $x$ are considered.

$\mathcal{U}(C_i)$ is the set of labels in $C_i$, $\min_{c_i}(\cdot)$ the element-wise minimum across distance maps per label and $\gamma C_i(x)$ a scaling and offset factor to increase the difference between adjacent labels.

$L_{reg}$ was chosen to be the l2-norm derivatives of the deformation ﬁeld~$T_i$:

\begin{equation}
	\label{eqn:regLoss}
	L_{reg}(T_i) = \sum_{\forall x \in T_i} \| \nabla u_i(x) \|_2^2.
\end{equation}

The volume preserving loss is based on \cite{dong2023preserving} and assures that the volume change at a given position is similar to the volume change of the surrounding organ:

\begin{equation}
	\label{eqn:volPresLoss0}
	L_{vol}(C_A,T_i) = \frac{1}{|X|} \sum_{x \in C_A} \operatorname{sigmoid}\left(5 \times (\mathcal{L}(x) - 1.5)\right),
\end{equation}

\begin{equation}
	\label{eqn:volPresLoss1}
	\mathcal{L}(x) = \max\left(\frac{|J(T_i)(x)|}{|\bar{J}(T_i, C_A(x))|} , \frac{|\bar{J}(T_i, C_A(x))|}{|J(T_i)(x)|} \right),
\end{equation}

\begin{equation}
	\label{eqn:volPresLoss2}
	\bar{J}(T_i, C_A(x)) = \frac{1}{|Y|} \sum_{y \in C_A = C_A(x)} J(T_i)(y).
\end{equation}

$(T_i)(x)$ is the determinant of the Jacobian matrix for transformation $T_i$ at position $x$ and $\bar{J}(T_i, C_A(x))$ in Equation \ref{eqn:volPresLoss2} is the mean Jacobian for the label $C_A(x)$ at position $x$.

\subsection{Handling Missing Correspondences}
\label{sec:missingCorrespondences}

\begin{figure*}[htb]
	\centering
	\resizebox{\textwidth}{!}{%
		\begin{tikzpicture}
			\def\dist{4.7cm}

			\node (leftTop) at (-\dist,0) {\includegraphics[width=0.49\linewidth]{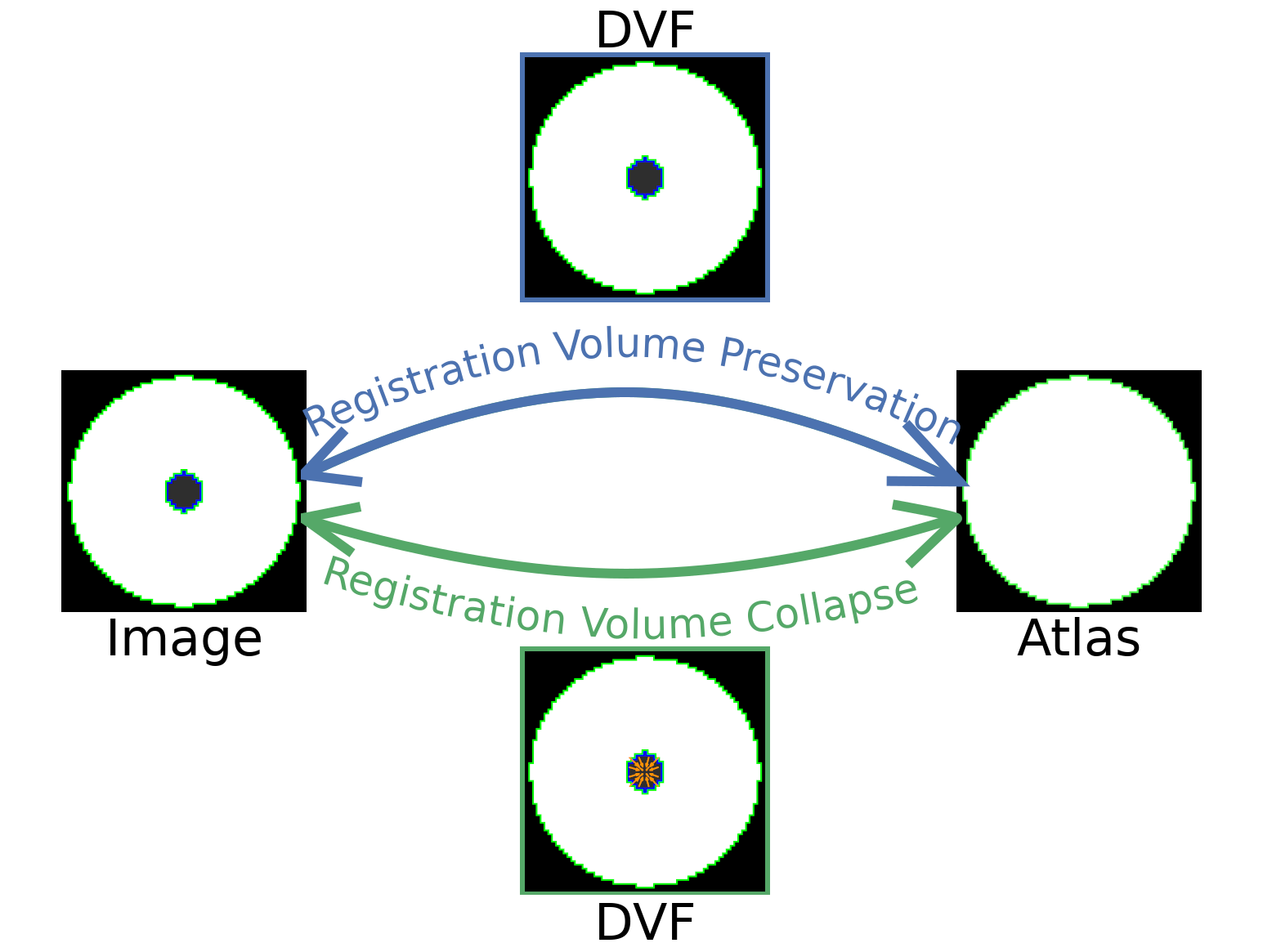}};
			\node (leftBottom) at (-\dist,-7cm) {\includegraphics[width=0.49\linewidth]{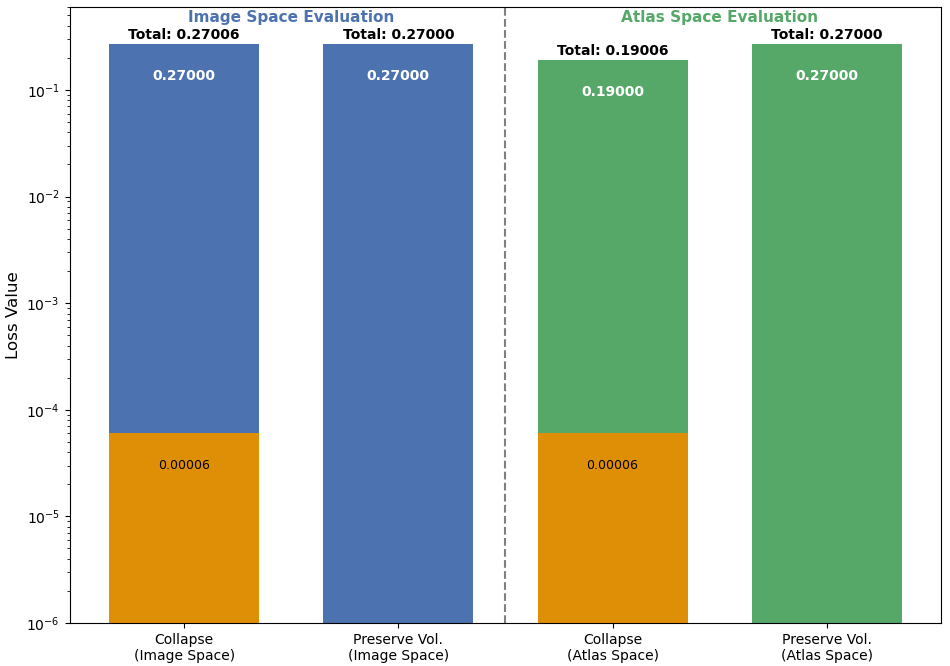}};

			\node (rightTop) at (\dist,0) {\includegraphics[width=0.49\linewidth]{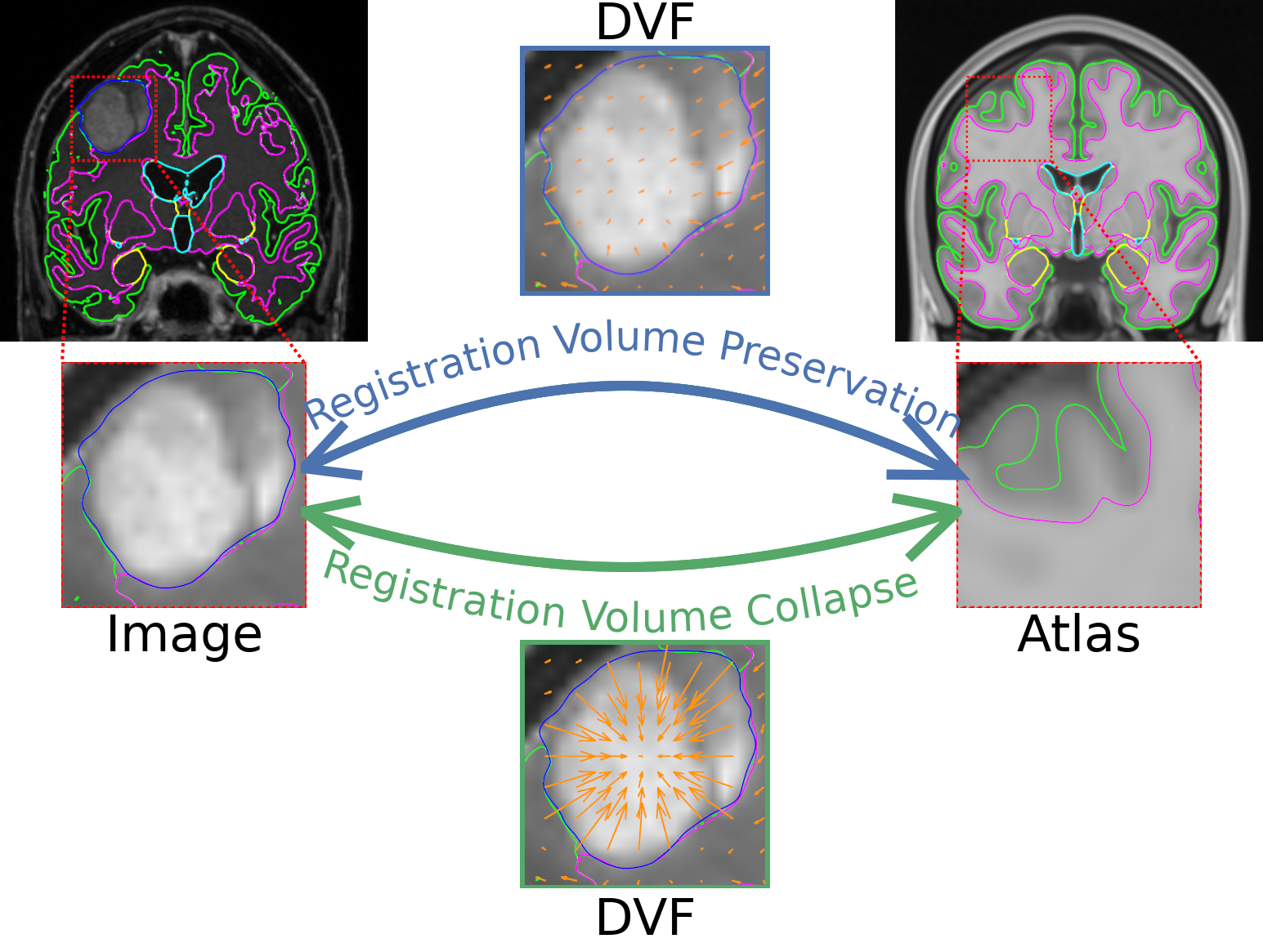}};
			\node (rightBottom) at (\dist,-7cm) {\includegraphics[width=0.49\linewidth]{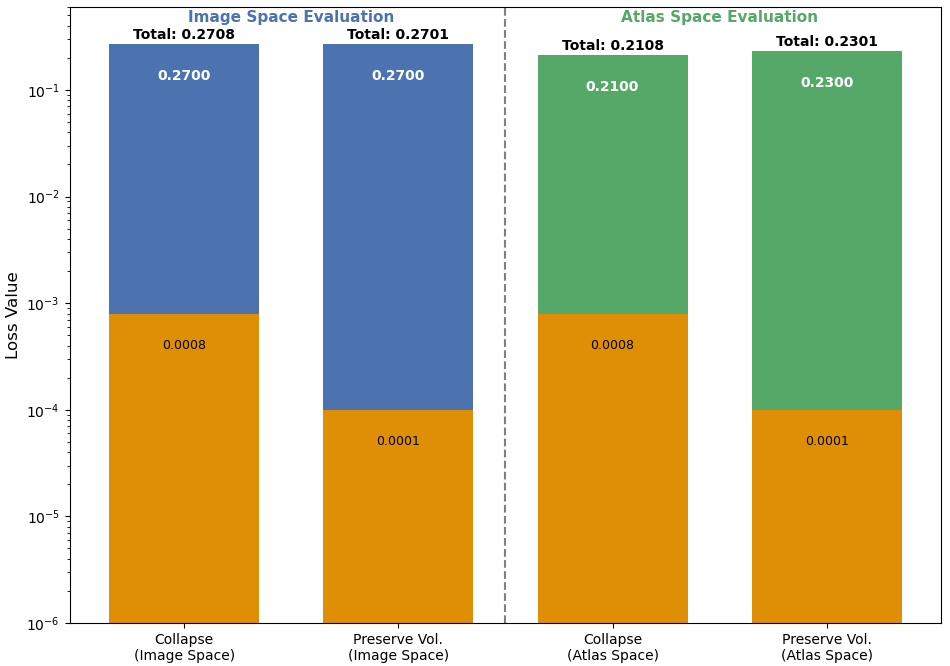}};

			\draw[thick] (0,3.3) -- (0,-9.8);
			\node at (0,-3.2) {\includegraphics[width=0.250\linewidth]{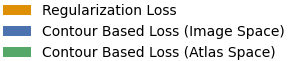}};
		\end{tikzpicture}
	}
	\caption{This figure illustrates the difference in contour-based loss between a collapsing tumour region and a volume-preserved tumour evaluated in image space and atlas space for a simplified (left) and a real world (right) example. The upper part of the figure shows the images with tumour contours (blue), the deformation vector fields (DVFs) in orange and the atlas. The lower part illustrates the calculated losses in image and atlas space for the DVF that collapses the tumour and the DVF that preserves the tumour volume. The plots demonstrate that an evaluation in atlas space favours the collapse of the tumour, while when evaluating the loss in image space, the volume preservation of the tumour is preferred.}
	\label{fig:collapseVisualization}
\end{figure*}

As noted earlier, a key challenge in registering two images arises when a structure present in one , such as a tumour, is absent in the other. These missing correspondences can degrade the quality of the deformation field. When training the over-fitting model by minimizing Equation~\ref{eqn:lossOverfitting}, we addressed this by introducing an additional loss term ($L_{vol}$). For the general model, this extra term is unnecessary because evaluating $L_{sim}$ directly in image space, as done in Equation~\ref{eqn:lossGeneral}, inherently penalizes tumour collapse: reducing the tumour volume leads to a higher loss than preserving it.
Figures~\ref{fig:collapseVisualization} and~\ref{fig:collapseVisualization2} illustrate this with two simple examples and one real-world case.

\subsection{Evaluation Metrics}

For the datasets used in this study, contours of brain structures and tumours were available.
To evaluate the registration algorithm, we computed the Sørensen--Dice index (\textit{DSC}),
Hausdorff distance (\textit{HD}), and average symmetric surface distance (\textit{ASSD}) before and after registration.
To detect anatomically implausible deformations, we calculated the determinant of the Jacobian matrix
for every point in a DVF and report the fraction of foldings (FoF) per image.
Tumour shrinkage is measured using the ratio of the tumour volume before and after registration.
Because deformation of the tissue surrounding the tumour also influences shrinkage,
we additionally report the ratio between the mean Jacobian determinant within the metastases and the mean Jacobian determinant
in the 10~mm peritumoral region.
As all analyses for this study were performed in atlas space,
we only provide metrics for registrations between individual datasets and the atlas.

\subsection{Experiments}

We began by assessing the registration algorithm’s accuracy on both open-source and clinical datasets.
Subsequently, we performed a clinical evaluation of all brain metastases.

\subsubsection{Registration Accuracy}
\label{sec:RegistrationAccuracy}

To establish a general baseline model and to ensure comparability with previously published methods, we trained the presented network on the publicly available Learn2Reg datasets.
Out of the 414 cases, 331 were used for training and 83 were used for testing.
The general model is further referred to as \textit{Learn2Reg-G}.

In the following, the general \textit{Learn2Reg-G} was over-fitted to each case in Learn2Reg and Clin1--3. This allows patient-specific characteristics to be taken into account and increases the quality of the registrations. The over-fitted models are further called \textit{Learn2Reg-OF}, \textit{Clin1-OF}, \textit{Clin2-OF}, and \textit{Clin3-OF}, respectively.

%

\subsubsection{Clinical Evaluation}

\paragraph{Metastases frequency per region}

The frequency with which MBM occur in different areas of the brain was investigated by transferring the tumour contour from each dataset to the respective atlas and examining in which atlas structure the tumour's centre of gravity lies. The measured frequencies per structure were then compared to the expected frequencies using the chi-square test. The expected frequencies were calculated by assuming a volume-corrected uniform distribution.
Bonferroni correction was used to counteract the multiple comparisons problem, and alpha was set to 1\%. One factor of uncertainty in this analysis is the accuracy of the registration. This was taken into account by randomly shifting each centre of gravity one hundred times by 1.0$\pm$0.5~mm. In this way, confidence intervals for the measured frequencies could be created. The value of 1.0$\pm$0.5~mm is a rough estimate of the registration error, which was determined by rounding up the ASSD reported in Section~\ref{sec:RegistrationAccuracy} to the nearest millimetre.

To assess whether MBM show a preferential localization close to the gray-white matter junction, as previously reported, we adapted the original method to account for the fjord-shaped nature of the gray-white matter junction and the resulting proximity of each point to this structure \cite{Barrios2025}. First, multiple sets of randomly sampled points across the brain were created. Then, the earth-mover distances between the set of tumour centres and the randomly sampled point sets
(\textit{$emd_{tumour}$}) and the distances among the random sets (\textit{$emd_{rand}$}) were calculated.
If the distances in $emd_{tumour}$ differ significantly from $emd_{rand}$, one could assume a higher frequency of MBM close to the gray-white matter junction. Perfusion characteristics were estimated by calculating the median, minimum, and maximum values for each transferred metastasis on the perfusion atlas.

%
\subsection{Implementation}

Hyperparameters (learning rate and $\lambda_{1-4}$) of the network were tuned with the Tune library \cite{liaw2018tune}. The following weighting factors were used for the loss functions (Equations~\ref{eqn:lossGeneral} and~\ref{eqn:lossOverfitting}) in all experiments: $\lambda_1$ = 0.098, $\lambda_3$ = 0.045, $\lambda_4$ = 0.098, and $\gamma$ = 1.0. The code and the Learn2Reg-G model are publicly available at: \url{https://github.com/ToFec/AtlasRegistration}

\section{Results}

All reported results were calculated in the ICBM152 atlas space. Unless otherwise stated, we report mean values $\pm$ standard deviation.

For the Learn2Reg dataset, the initial values after affine pre-registration for DSC, HD, and ASSD were 0.59$\pm$0.22, 8.12$\pm$5.69~mm, and 1.49$\pm$0.73~mm, respectively. The Learn2Reg-G model achieved a DSC of 0.75$\pm$0.20, a HD of 8.02$\pm$6.37~mm, and an ASSD of 0.91$\pm$0.41~mm on the independent Learn2Reg test set. The moderate DSC performance can be attributed to two structures:
the left and right inferior lateral ventricles, as well as the left and right generic blood vessel label. Excluding those two thin and small structures raises the DSC to 0.80$\pm$0.12.
Learn2Reg-OF achieved a DSC of 0.92$\pm$0.09, a HD of 6.79$\pm$5.70~mm, and an ASSD of 0.63$\pm$0.23~mm on SegAnat.

On average, Clin1, Clin2, and Clin3 had an initial registration error after affine pre-registration with a DSC of 0.61$\pm$0.21, a HD of 8.32$\pm$6.73~mm, and an ASSD of 1.47$\pm$1.21~mm. The respective OF-models improved the results to a DSC of 0.89$\pm$0.12, a HD of 7.23$\pm$6.39~mm, and an ASSD of 0.76$\pm$0.38~mm. The tumour volume decreased on average by a factor of 0.84$\pm$0.38,
and the average ratio of the mean Jacobian determinant of the tumour to its 10~mm surrounding was 0.93$\pm$0.37.
Detailed results can be found in Table~\ref{tab:registrationResults}.

%
%

To measure the impact of the volume preserving loss, we retrained Clin1-OF with $\lambda_4$ set to zero. While this did not change the average DSC, tumour volume decreased by a factor of 0.73$\pm$0.49 and the Jacobian determinant ratio was found to be 0.85$\pm$0.45.
These differences were statistically significant.

Analysis of the DVF with respect to anatomically implausible deformations showed that the FoF was, on average, 0.3\% of the calculated deformation vectors across the image datasets.

\begin{table}[h!]
	\centering
	\caption{Performance metrics of all trained models on the respective datasets. The first part of the Dataset/Model name indicates the dataset,
		the second part stands for the model type (A: affine, G: general, OF: over-fitted). Column two through four give the Dice coefficient (DSC), Hausdorff distance (HD), and average surface distance (ASSD). The tumour volume factor gives the ratio of the tumour volume before (affine) and after deformable (OF) registration.
		The average ratio of the mean Jacobian determinant of the tumour and its 10~mm surrounding is given in the column labeled Jacobian ratio. Values are presented as mean$\pm$standard deviation.
	}
	\label{tab:registrationResults}
	\begin{tabular}{lcccccc}
		\hline
		\textbf{Dataset/Model} & \textbf{DSC}  & \textbf{HD [mm]} & \textbf{ASSD [mm]} & \makecell{\textbf{Tumour}                 \\ \textbf{Volume Factor}} & \makecell{\textbf{Jacobian} \\ \textbf{Ratio}} \\
		\hline
		Learn2Reg-A            & 0.59$\pm$0.22 & 8.12$\pm$5.69    & 1.49$\pm$0.73      & --                        & --            \\
		Learn2Reg-G            & 0.75$\pm$0.20 & 8.02$\pm$6.37    & 0.91$\pm$0.41      & --                        & --            \\
		Learn2Reg-OF           & 0.92$\pm$0.09 & 6.79$\pm$5.70    & 0.63$\pm$0.23      & --                        & --            \\
		\hline
		Clin1-A                & 0.61$\pm$0.21 & 8.58$\pm$7.11    & 1.57$\pm$1.56      & --                        & --            \\
		Clin1-OF               & 0.89$\pm$0.12 & 7.27$\pm$6.51    & 0.77$\pm$0.39      & 0.82$\pm$0.38             & 0.91$\pm$0.41 \\
		\hline
		Clin2-A                & 0.61$\pm$0.21 & 7.92$\pm$6.15    & 1.34$\pm$0.56      & --                        & --            \\
		Clin2-OF               & 0.90$\pm$0.10 & 7.09$\pm$6.14    & 0.74$\pm$0.36      & 0.85$\pm$0.40             & 0.94$\pm$0.31 \\
		\hline
		Clin3-A                & 0.58$\pm$0.22 & 8.42$\pm$6.61    & 1.47$\pm$0.67      & --                        & --            \\
		Clin3-OF               & 0.89$\pm$0.13 & 7.60$\pm$6.75    & 0.77$\pm$0.40      & 0.86$\pm$0.25             & 0.98$\pm$0.33 \\
		\hline
	\end{tabular}
\end{table}

\subsubsection{Clinical Evaluation}

We could not detect any differences in the frequency of MBM between the right and left hemispheres. Therefore, no distinction was made between the two hemispheres in the subsequent analysis.

The anatomical evaluation of SegAnat revealed that most metastases occurred in the cerebral white matter (131) and the cerebral cortex (309), followed by the cerebellar cortex (36) and the putamen (20). However, when the expected frequencies are considered, it becomes apparent that metastases in the cerebral white matter are highly under-represented (131 vs.\ 194), while metastases in the cerebral cortex (309 vs.\ 224) and the putamen (20 vs.\ 5) are over-represented. All three differences were statistically significant. No significant association between metastasis volume and anatomical localization was observed, and we did not detect significantly smaller metastases in the putamen, differing from observations reported by Lasocki et al. \cite{Lasocki2019}.

\begin{figure*}[htbp]
	\centering
	\mbox{
		\shortstack{
			\includegraphics[width=0.49\linewidth]{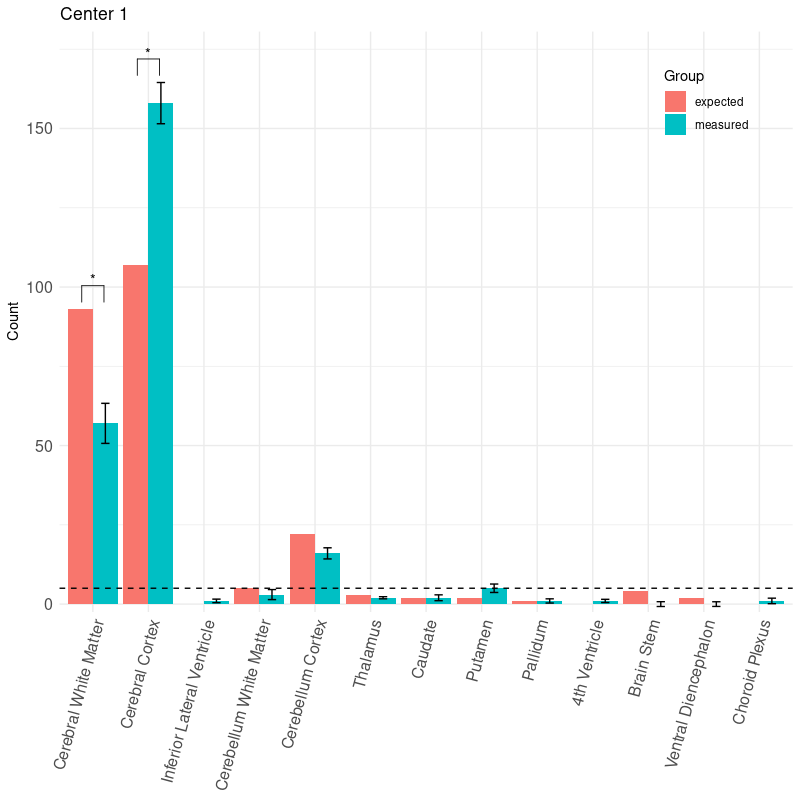}
		}
		\shortstack{
			\includegraphics[width=0.49\linewidth]{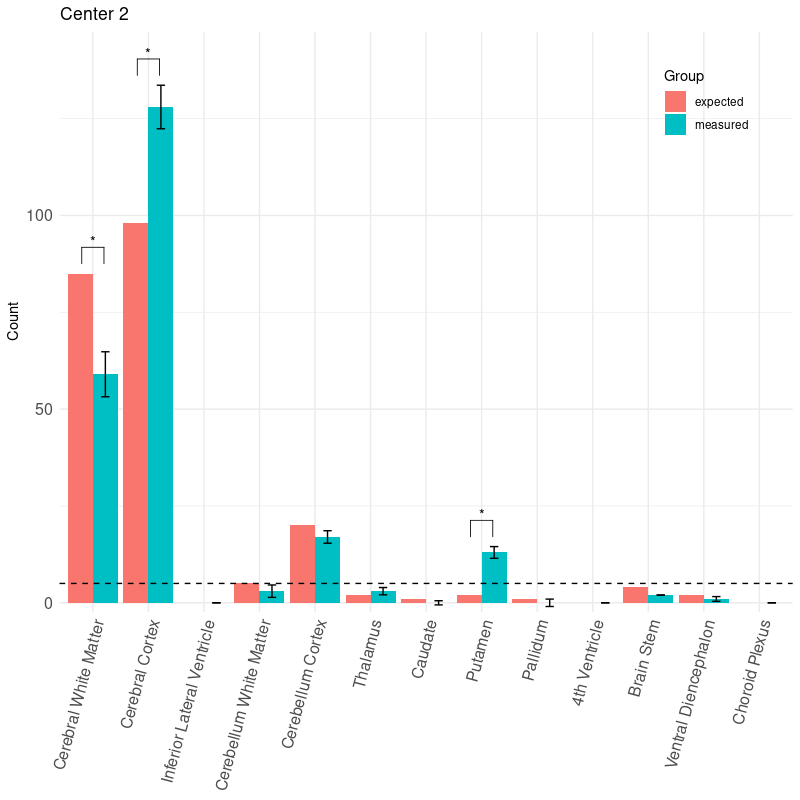}
		}
	}
	\mbox{
		\shortstack{
			\includegraphics[width=0.49\linewidth]{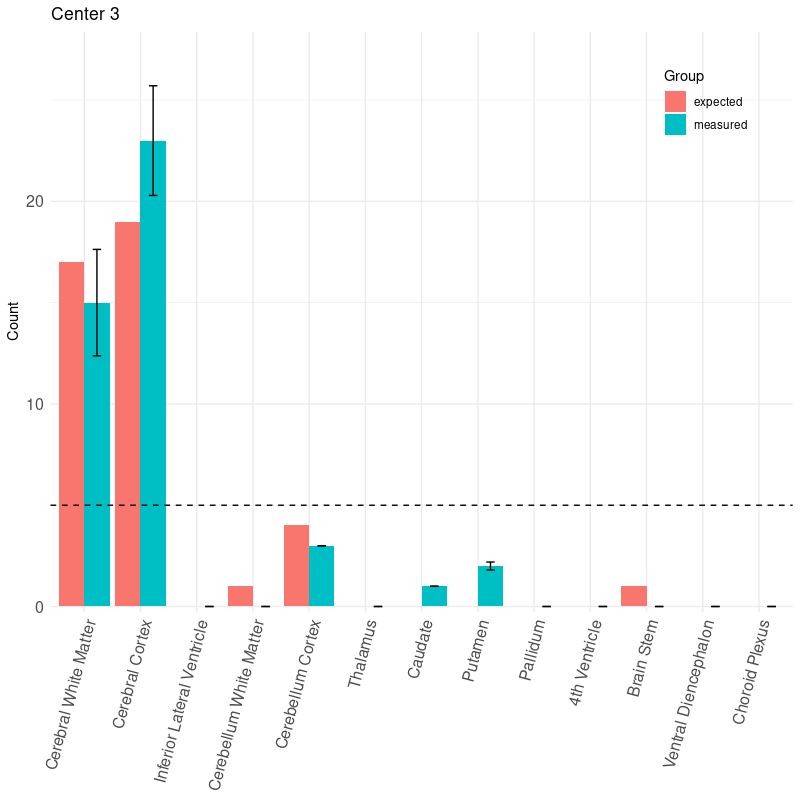}
		}
		\shortstack{
			\includegraphics[width=0.49\linewidth]{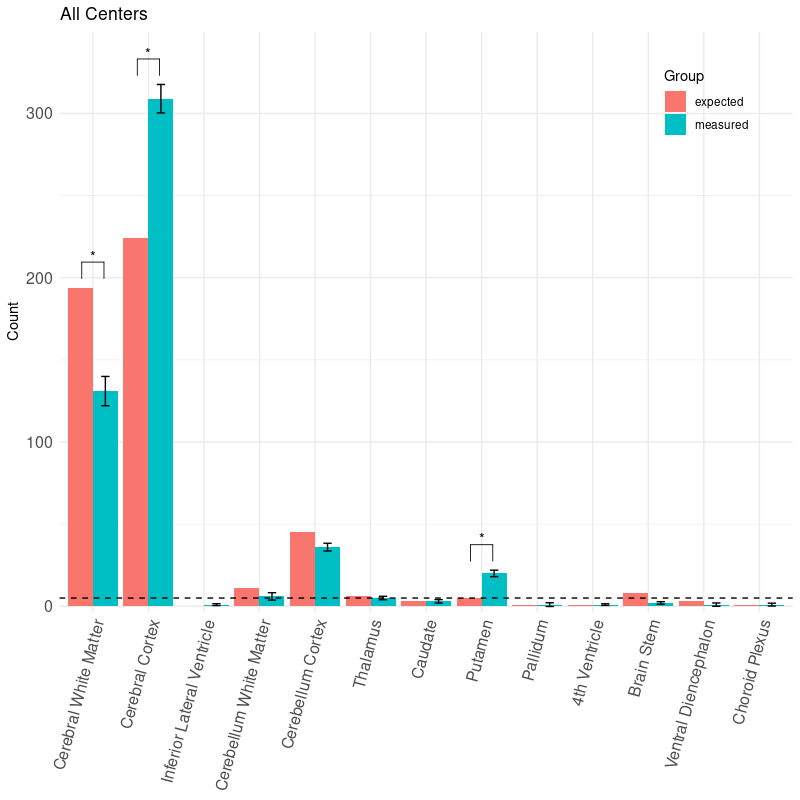}
		}
	}
	\caption{The expected (green) number of occurrences (assuming a uniform distribution) versus the measured (red) occurrences for all three centres and a combination of all three datasets. The stars indicate statistically significant differences, and the dashed line indicates the minimum number of samples required for the chi-square test. The figure shows counts only for brain areas with at least one metastasis in at least one of the three centres.
	}
	\label{fig:regionCount}
\end{figure*}

\begin{figure*}[htbp]
	\centering
	\mbox{
		\includegraphics[width=0.8\linewidth]{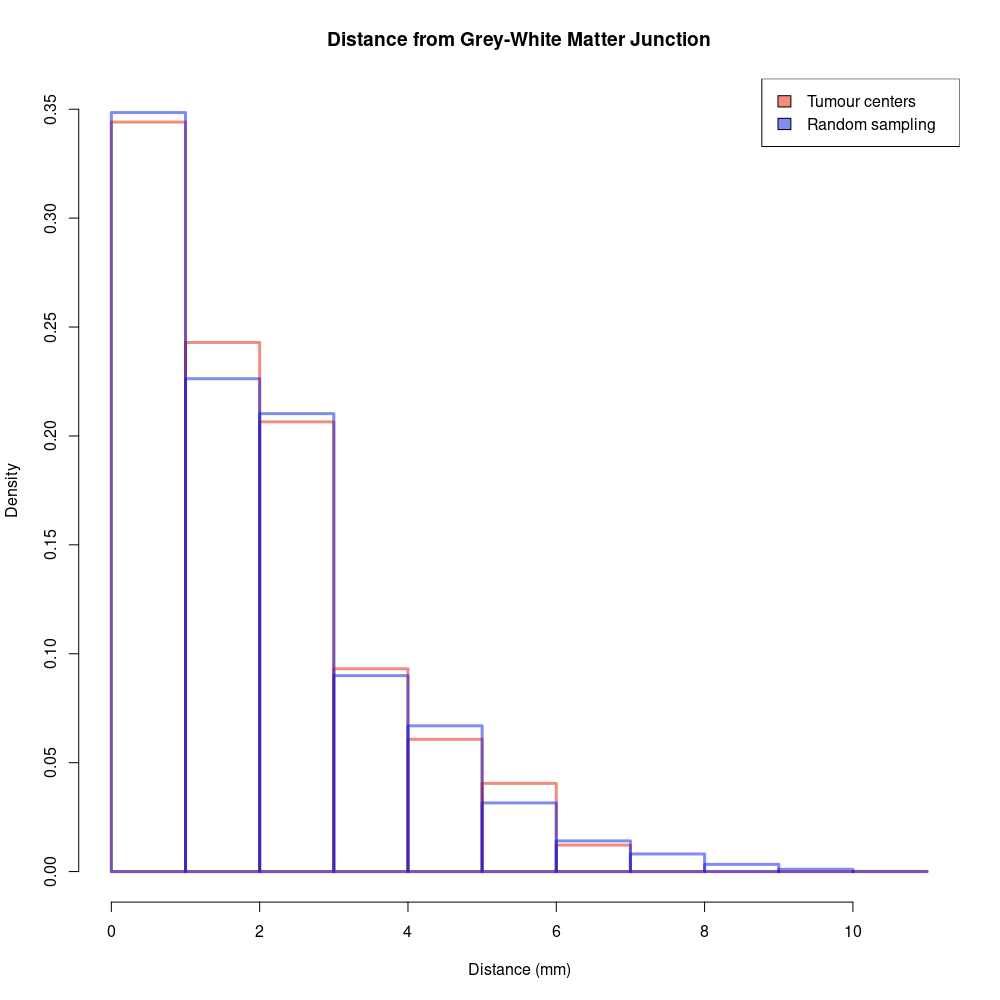}
	}
	\caption{Histogram of distance to the gray-white matter junction for melanoma metastasis barycentres and randomly sampled points. The y-axis represents the density of occurrences, while the x-axis shows the distance to the gray-white matter junction. The histogram includes two curves: one for metastasis barycentres (red line) and one for randomly sampled points (blue line). Compared to random sampling, metastasis barycentres exhibit a significantly higher density within a 1–2~mm distance from the gray-white matter junction, indicating a preferential spatial distribution of melanoma metastases in this region.}
	\label{fig:greyWhiteMatterJunction}
\end{figure*}

For the cerebellar cortex, a trend toward under-representation was observed (36 vs.\ 45), although this did not reach statistical significance. The remaining structures did not reach the count necessary for statistical testing.
A detailed graphical analysis for each centre is given in Figure~\ref{fig:regionCount},
and the corresponding numbers are provided in Table~\ref{tab:metastasis_counts}.
The histogram in Figure~\ref{fig:greyWhiteMatterJunction} illustrates the distribution of distances from the
gray-white matter junction to MBM barycentres and randomly sampled points.
Tumour centres were significantly ($p < 0.001$) more concentrated near the junction,
suggesting a non-random spatial preference in this region compared to random sampling. 
In addition, tumour centres showed a significant spatial preference for the corticomeningeal interface, although this effect was less pronounced than for the gray–white matter junction.

Analyses based on the perfusion atlas and the arterial territory atlas did not reveal statistically significant spatial preferences. Metastases from all centres were found in areas with similar normalized perfusion.
The average median values were 0.57, 0.55, and 0.58 for centres~1, 2, and~3, respectively. The corresponding minimum and maximum values were on average 0.45, 0.43, and 0.47, and 0.67, 0.65, and 0.68, respectively. Similarly, no arterial territory was associated with an increased occurrence of MBM. This applies to both the detailed subdivision into arterial areas of influence
(listed in Table~\ref{tab:structureSetDetails})
and when these areas are summarized by distinguishing only between anterior and posterior territories.

\begin{table}[htbp]
	\centering
	\caption{Measured and expected metastasis counts per brain region across three centres and the combined dataset. Expected counts were calculated by assuming a volume-corrected uniform distribution.}
	\setlength{\tabcolsep}{3pt}
	\label{tab:metastasis_counts}
	\resizebox{\linewidth}{!}{
		\begin{tabular}{l
				cc 
				cc 
				cc 
				cc 
			}
			\hline
			\multirow{2}*{\textbf{Region}}        &
			\multicolumn{2}{c}{\textbf{Centre 1}} &
			\multicolumn{2}{c}{\textbf{Centre 2}} &
			\multicolumn{2}{c}{\textbf{Centre 3}} &
			\multicolumn{2}{c}{\textbf{Combined}}                                                                          \\[2pt]
			                                      & \textbf{Measured} & \textbf{Expected} &
			\textbf{Measured}                     & \textbf{Expected} &
			\textbf{Measured}                     & \textbf{Expected} &
			\textbf{Measured}                     & \textbf{Expected}                                                      \\
			\hline
			Cerebral White Matter                 & 57                & 93                & 59  & 85 & 15 & 17 & 131 & 194 \\
			Cerebral Cortex                       & 158               & 107               & 128 & 98 & 23 & 19 & 309 & 224 \\
			Lateral Ventricle                     & 0                 & 4                 & 0   & 3  & 0  & 1  & 0   & 7   \\
			Inferior Lateral Ventricle            & 1                 & 0                 & 0   & 0  & 0  & 0  & 1   & 0   \\
			Cerebellum White Matter               & 3                 & 5                 & 3   & 5  & 0  & 1  & 6   & 11  \\
			Cerebellum Cortex                     & 16                & 22                & 17  & 20 & 3  & 4  & 36  & 45  \\
			Thalamus                              & 2                 & 3                 & 3   & 2  & 0  & 0  & 5   & 6   \\
			Caudate                               & 2                 & 2                 & 0   & 1  & 1  & 0  & 3   & 3   \\
			Putamen                               & 5                 & 2                 & 13  & 2  & 2  & 0  & 20  & 5   \\
			Pallidum                              & 1                 & 1                 & 0   & 1  & 0  & 0  & 1   & 1   \\
			3rd Ventricle                         & 0                 & 0                 & 0   & 0  & 0  & 0  & 0   & 0   \\
			4th Ventricle                         & 1                 & 0                 & 0   & 0  & 0  & 0  & 1   & 1   \\
			Brain Stem                            & 0                 & 4                 & 2   & 4  & 0  & 1  & 2   & 8   \\
			Hippocampus                           & 0                 & 2                 & 0   & 2  & 0  & 0  & 0   & 4   \\
			Amygdala                              & 0                 & 1                 & 0   & 1  & 0  & 0  & 0   & 1   \\
			Accumbens Area                        & 0                 & 0                 & 0   & 0  & 0  & 0  & 0   & 1   \\
			Ventral Diencephalon                  & 0                 & 2                 & 1   & 2  & 0  & 0  & 1   & 3   \\
			Vessel                                & 0                 & 0                 & 0   & 0  & 0  & 0  & 0   & 0   \\
			Choroid Plexus                        & 1                 & 0                 & 0   & 0  & 0  & 0  & 1   & 1   \\
			\hline
		\end{tabular}
	}
\end{table}

\section{Discussion}
In this work, we presented a deep learning-based deformable atlas-registration framework. The method is specifically designed to address key challenges in atlas registration arising from focal pathological lesions that introduce missing correspondences and thereby enables systematic, reproducible spatial analyses across large, heterogeneous clinical cohorts.
By applying the framework to data from three independent centres,
we demonstrated high registration accuracy and showed that the proposed approach facilitates the extraction of meaningful neuroanatomical insights into the spatial distribution of MBM.

A key strength of the proposed framework is its sampling-module-based image handling,
which eliminates the need for conventional preprocessing steps such as resampling and interpolation. This approach preserves the original image information and substantially reduces image degradation that would otherwise accumulate through repeated interpolation operations.

A second major strength is the method’s robust handling of missing correspondences,
achieved without the need for lesion masks, tuned thresholds, or pre-trained auxiliary components.
The forward-mode similarity evaluation on distance-transformed label maps implicitly penalizes unrealistic collapses of lesion volumes, while the one-shot over-fitting stage provides additional stabilization via a volume-preserving loss. Taken together, these components enable anatomically plausible deformations even in cases with large or irregular metastases.

The registration results further support the effectiveness of the proposed framework.
Across all clinical datasets, the over-fitted models achieved consistently high DSC values (0.89--0.90)
and a consistent reduction in average surface distances.
Tumour shrinkage, when present, was limited in magnitude,
and the Jacobian-based analysis indicated that deformation within metastatic regions was comparable to that of their local surroundings,
thereby supporting the anatomical plausibility of the registrations.

While performance on the Learn2Reg benchmark was moderate to good for the general model,
over-fitting led to substantial improvements in registration accuracy, which is consistent with recent findings that one-shot tuning can considerably enhance patient-specific alignment. However, further improvements remain possible.
For example, the loss function could be adapted to place greater emphasis on small or thin anatomical structures,
or more powerful network architectures could be incorporated in the framework.


From a methodological perspective, melanoma brain metastases constitute a clinically relevant and biologically distinct subgroup of brain metastases. Focusing on metastases from a single primary tumour enabled a direct comparison with previously published melanoma-specific studies, thereby supporting interpretation of the observed spatial patterns as both a validation of existing literature and an assessment of the framework’s ability to reproduce known biological effects. Importantly, this approach avoids confounding introduced by pooling metastases from different primary entities, which are known to exhibit distinct intracranial localization patterns \cite{Neman2022,Mahmoodifar2024}.

Our results both corroborate and refine previously reported observations. Consistent with previous studies, MBM were primarily located in the cerebral cortex and cerebral white matter,
while our volume-adjusted analysis revealed that metastases were significantly over-represented in the cortex and under-represented in white matter. We did not observe a statistically significant under-representation in the cerebellum, as previously reported \cite{Mampre2019,Schroeder2020},
although a non-significant trend toward fewer lesions was observed. The analysis across arterial territories yielded no preferential distribution in either anterior or posterior circulations, which contrasts with findings from larger multi-primary datasets that suggested a stronger anterior-circulation predominance, especially within regions supplied by the middle cerebral artery \cite{Lyu2025}.

The putamen showed a significant over-representation, for which previous studies have reported heterogeneous findings regarding the involvement of deep brain structures. Lyu et al. observed a sparing of gray matter, whereas Lasocki et al. showed that especially smaller MBM were more likely to be located in lenticulostriate vascular territory \cite{Lyu2025,Lasocki2019}. The preferential involvement of the putamen could reflect a combination of its dense lenticulostriate vascular supply, its relatively large volume among basal ganglia structures, and the increased sensitivity afforded by volume-corrected atlas-based analyses.

A notable confirmation of earlier large-cohort studies is the preferential proximity of melanoma brain metastases to the gray-white matter junction \cite{Hwang1996,Barrios2025}. Our additional observation of increased localization toward the corticomeningeal interface is consistent with findings by Lasocki et al., who reported that small melanoma brain metastases frequently develop in close association with the leptomeninges \cite{Lasocki2019}.
The preferential localization of melanoma brain metastases at cortical border zones, including the gray-white matter junction and the corticomeningeal interface, is biologically plausible and likely reflects vascular and hemodynamic factors. These regions are characterized by abrupt changes in vessel caliber and reduced flow velocities, which increase the likelihood of mechanical trapping and extravasation of circulating tumour cells \cite{Kienast2010,Valiente2018,Boire2020}.

Despite these strengths, technical and clinical limitations should be acknowledged. From a technical perspective, issues regarding the proposed registration approach need to be considered.
First, although our sampling-module approach avoids traditional preprocessing, the method relies on an initial affine alignment step.
Suboptimal affine initialization may affect the quality of subsequent deformable registration. Second, while distance-transformed label maps provide robust shape-based similarity, they depend on the accuracy of the underlying image segmentation. Inaccuracies in these anatomical labels may therefore propagate into the registration process. Finally, although over-fitting substantially improves registration accuracy,
it comes at the cost of increased computational demand and may lead to over-specialization to local noisy image characteristics if not adequately regularized.

In addition to these technical considerations, certain clinical and study-related limitations inherent to the analyzed data should be considered. Although our three clinical datasets were heterogeneous in acquisition and scanner parameters - thereby demonstrating robustness of the framework -, inter-centre variability in tumour delineation may introduce systematic bias in the spatial analysis of metastasis locations. Furthermore, the number of cases in some brain structures was low, limiting statistical power and precluding robust conclusions for several smaller or less frequently affected subcortical regions.

The proposed atlas-based framework maps individual MBM into a common anatomical reference space, enabling automated and reproducible large-scale analyses. This standardized spatial representation is of particular value for systematic cohort-level investigations of brain metastases, such as the identification of anatomical regions that may be associated with an increased risk of local recurrence or treatment-related toxicity. Future work will focus on improving the performance of the general model, reducing its dependency on anatomical labels, and integrating uncertainty measures for deformation fields. The availability of the proposed framework as an open-source tool supports transparent reuse and facilitates independent validation and extension.

\section{Conclusion}
\label{sec:conclusion}
We have presented a fully differentiable, deep-learning-based registration framework that (i) operates on native medical images without the need for extensive preprocessing, (ii) robustly handles missing correspondences caused by pathological lesions,
and (iii) provides a publicly available implementation. Its application to three independent MBM cohorts demonstrated high registration performance and enabled a systematic spatial characterization of metastasis distribution in a common anatomical space.

Our findings corroborate a preferential localization of MBM near the gray-white matter junction
and demonstrate significant over-representation in the cerebral cortex and putamen,
while white matter is comparatively under-represented.
These results extend and refine existing knowledge derived from previously heterogeneous imaging cohorts
and highlight the value of standardized atlas-based analyses.

By publicly releasing our framework, we aim to provide a practical and robust tool that can support future neuroanatomical, radiomics, and clinical studies involving pathological brain images. This work demonstrates the potential of advanced registration techniques to improve the consistency, reproducibility, and biological interpretability of multi-centre neuro-oncological imaging research.

\section*{Data Statement}
The deep learning registration framework, including the source code and the Learn2Reg-G model weights, is publicly available on GitHub (URL: \url{https://github.com/ToFec/AtlasRegistration}).

\section*{Ethics Statement}
This multicentre retrospective study was conducted in accordance with the Declaration of Helsinki. Primary ethical approval was obtained from the local ethics committee of the Medical Faculty of the University of Freiburg (institutional review board approval number 20-1031\_3). Additional approvals were secured from the ethics committees of the other participating centres.

\section*{Declaration of Competing Interest}
The authors declare that they have no known competing financial interests or personal relationships that could have appeared to influence the work reported in this paper.

\section*{Author contributions: CRediT}
\textbf{Nanna E. Wielenberg:} Data curation, Investigation, Formal analysis, Writing – original draft. \textbf{Ilinca Popp:} Data curation, Writing – review and editing. \textbf{Oliver Blanck:} Data curation, Writing – review and editing. \textbf{Lucas Zander:} Data curation, Writing – review and editing. \textbf{Jan C. Peeken:} Data curation, Writing – review and editing. \textbf{Stephanie E. Combs:} Data curation, Writing – review and editing.
\textbf{Anca-Ligia Grosu:} Supervision, Writing – review and editing. \textbf{Dimos Baltas:} Supervision, Writing – review and editing. \textbf{Tobias Fechter:} Conceptualization, Investigation, Methodology, Formal analysis, Writing – original draft.

\bibliographystyle{elsarticle-num}
\bibliography{bibliography}
\newpage
\appendix
\setcounter{figure}{0}
\renewcommand{\thefigure}{A.\arabic{figure}}

\section{}

\begin{figure*}[htb]
	\centering
	\mbox{
		\includegraphics[width=0.8\linewidth]{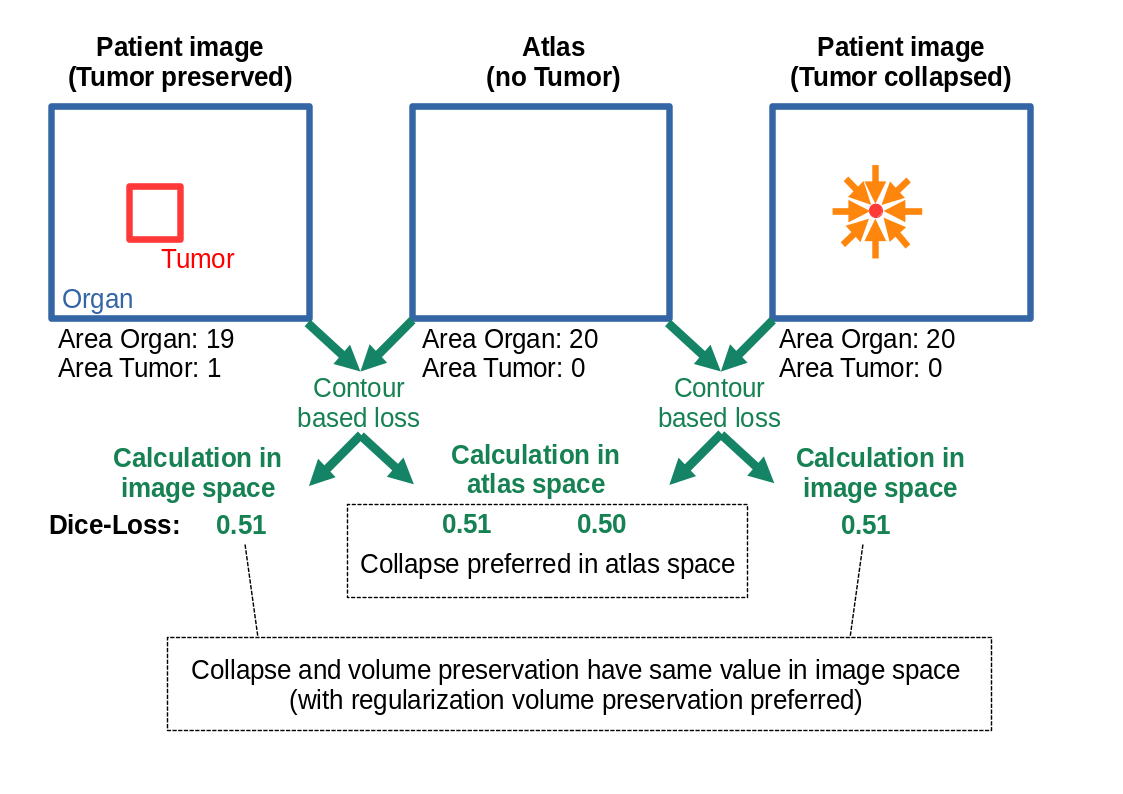}
	}
	\caption{A simplified example illustrates that a collapse of the tumour region
		reduces the loss only when it is evaluated in atlas space, but not in image space.
		The orange arrows indicate the deformation vectors.
		When the tumour is preserved (left), no deformation is required.
		Due to a more irregular deformation field and therefore a higher regularization loss,
		the collapse of the tumour region results in a higher loss in image space compared to tumour preservation.}
	\label{fig:collapseVisualization2}
\end{figure*}

\end{document}